\documentclass[twoside,11pt]{article}
\usepackage{jmlr2e}
\usepackage{graphicx}
\usepackage{subcaption}

\usepackage{xcolor}

\ShortHeadings{Hallucinations or Attention Misdirection? }{Ioste}
\firstpageno{1}

\begin{document}

\title{Hallucinations or Attention Misdirection? The Path to Strategic Value Extraction in Business Using Large Language Models}

\author{\name Aline Ioste \email ioste@ime.usp.br \\
       \addr Institute of Mathematics and Statistics\,
       University of São Paulo (IME-USP)\\
       \addr Artificial Intelligence Specialist in the Banking Sector - Brazil}

\editor{}

\maketitle

\begin{abstract}
Large Language Models (LLMs) with transformer architecture have revolutionized the domain of text generation, setting unprecedented benchmarks. Despite their impressive capabilities, LLMs have been criticized for generating outcomes that deviate from factual accuracy or display logical inconsistencies, phenomena commonly referred to as “hallucinations”. This term, however, has often been misapplied to any results deviating from the instructor's expectations, which this paper defines as attention misdirection rather than true hallucinations. Understanding the distinction between hallucinations and attention misdirection becomes increasingly relevant in business contexts, where the ramifications of such errors can significantly impact the value extraction from these inherently pre-trained models. This paper shows the best practices of the PGI method, a strategic framework that achieved a remarkable error rate of only 3.15\% in 4,000 responses generated by GPT to a real business challenge, highlighting that by equipping experimentation with knowledge, businesses can unlock opportunities for innovate through the use of these natively pre-trained models. This reinforces the notion that strategic application grounded in a skilled team can maximize the benefits of emergent technologies such as the LLMs.

\end{abstract}

\begin{keywords}
  Hallucinations, Generative Artificial Intelligence, Attention Dynamics, Large Language Model with Transformers Architecture, Business problem
\end{keywords}

\section{Introduction}

Large Language Models (LLMs) with transformer architecture have revolutionized natural language processing \citep{vaswani2017attention}, enabling machines to generate human-like text with unprecedented accuracy and fluency. However, amid the remarkable progress, a curious phenomenon has emerged: hallucinations within the outputs of these models \citep{ji2023survey}.

Hallucinations, traditionally associated with human perception \citep{lieberman2014hallucination}, manifest as instances where LLMs produce text that diverges from factual reality or exhibits inconsistencies that defy logical coherence \citep{ji2023survey}. This intriguing occurrence raises fundamental questions about the inner workings and limitations of LLMs, prompting researchers to scrutinize the underlying mechanisms driving such anomalies \citep{ji2023survey}.

The phenomenon of hallucinations in LLMs is well-documented in academic research, highlighting instances where models produce factually incorrect or logically inconsistent information \citep{rebuffel2022controlling}. The academic research aims to understand and mitigate such errors, enhancing LLM reliability and comprehension \citep{ji2023survey}. Conversely, a broader range of users, including professionals across various sectors, often misapplies the term "hallucination" to any model output that fails to meet their expectations, regardless of its factual accuracy. 

This paper delves into the phenomena often described as hallucinations in LLMs, proposing that many such instances are more accurately characterized as errors of attentional misdirection rather than authentic hallucinatory experiences. Through meticulous observation and analysis of projects executed by data scientists and engineers, a pronounced knowledge disparity regarding these concepts has been identified. This insight, gained from direct and indirect engagement with various LLM initiatives, emphasizes the critical need for enhanced clarity and precision in communication within professional environments. 

The prevalent misuse of terminology not only reveals a considerable gap in the understanding of LLMs among practitioners but also mirrors a wider misunderstanding regarding the operational strengths and limitations of these models. 

This paper clarifies the adoption of precise terms in professional contexts to effectively exploit the advanced potential of AI technologies such as LLMs. Shows how the PGI framework \citep{2023arXiv230813317I} emphasizes strategic design, linguistic conceit, and contextual clarity to guide LLMs toward accurate, business-relevant answers. This strategic approach not only reduces the risk of misdirected attention but also leverages the models' knowledge base, demonstrating the importance of precise communication and understanding in human-machine interactions for innovative solutions and value creation in the business landscape.

\subsection{The Term "Hallucinations" can be potentially misleading}

Applying the term "hallucinations" to the outputs of LLMs should be done sparingly as its widespread use can be misleading. It suggests a direct connection to human perceptual phenomena, which are inherently subjective and often linked to psychiatric or neurological disorders \citep{macpherson2013hallucination}. 

This use risks anthropomorphizing language models, attributing them human-like perceptual qualities, whereas LLMs are computational systems driven by algorithms and data, lacking human consciousness or perception \citep{liu2024survey}. Moreover, the term has been inaccurately deployed by many professionals to describe attention errors, often due to prompt mistakes, rather than true "hallucinations" defined scientifically as outputs that are factually incorrect or logically inconsistent. 

Labeling any output that does not meet the instructor's expectations as "hallucinations" can misconstrue these phenomena as inevitable or uncontrollable, obscuring the understanding of underlying mechanisms and hindering the development of effective mitigation strategies  \citep{guerreiro2023hallucinations}. Therefore, it's critical to avoid the generalized use of "hallucinations" in analyzing LLM outcomes to promote a clearer comprehension of their functionalities and to facilitate the development of more accurate approaches for managing outputs that do not align with intended objectives.

This paper distinguishes between two types of errors in the outputs of LLMs: 1ª: human-induced attention misdirection, and 2ª: inherent model flaws, factually incorrect, or logically inconsistent responses. This paper specifically clarifies that the first category, attentional misdirection errors, are not genuine hallucinations and can be mitigated strategically.

\section{Importance of Linguistic Skills and Language Understanding in Efficient Human-Machine Interaction}

LLMs with transformers have significantly advanced the understanding of semantics in language models \citep{vaswani2017attention}. Unlike earlier models that mainly focused on statistical patterns of language \citep{manaswi2018rnn}. The linguistic study was essential to design and train models effectively, as early language models relied heavily on statistical methods and features derived from linguistic insights \citep{khan2019rnn}. With the development of transformers, the focus shifted towards models learning directly from large datasets, reducing the need for explicit linguistic knowledge in the training process \citep{vaswani2017attention}. Transformers learn patterns and relationships within the data, automating much of what previously required strongly linguistic analysis \citep{xu2023multimodal}.

They do this by utilizing attention mechanisms, which allow the model to weigh the importance of different words in a sentence or document based on their context. The capacity to comprehend the context and the interplay between words empowers transformers to understand subtle nuances of meaning, rendering them exceptionally proficient for tasks demanding a comprehension of language semantics \citep{ghojogh2020attention}.

However, this technological leap has led many professionals to underestimate the value of linguistic studies and to operate at a superficial level of language structure, often based on unfounded immediate tips, not grounded in science or linguistic research. Many have inefficiently tried to extract value from these models without a deep understanding of the linguistic principles of natural language.

Drawing an analogy with programming languages like Python, Java, or C++, which, despite being designed to be intuitive, do not eliminate the need for deep study to avoid syntax errors and fully leverage their potential \citep{schmidt1996programming}. This analogy illustrates that understanding the syntax and fundamental principles for efficient coding requires grasping the structure of the language. Similarly, instructing a machine using a more intuitive language also necessitates a deep understanding of linguistic nuances to prevent machine misinterpretations.

The study of the language used to direct a machine, whether it be programming language or natural language, remains crucial. Combining the potential of transformers with linguistic insights can significantly enhance the understanding and application of the model in complex semantic tasks.

To derive business value and steer these models toward a specific business task requires human agent awareness responsible for creating these directions (prompts). In communicating with a machine through natural language, it is imperative to consider its capacity to process this information and ensure the instructions comply with the rules and syntax of the used language \citep{acquaviva2022communicating}.

It is crucial to remember not to anthropomorphize the machine, as it has no awareness of the input or output data; interpreting this information is the exclusive responsibility of the only conscious being in this interaction, humans \citep{proudfoot2011anthropomorphism}. Thus, any feedback not aligned with the human instructor's expectation should be carefully considered concerning the given instruction before characterizing it as a hallucination of the machine.

To ensure effective interaction, maintaining vigilance over the specific syntaxes of the language is fundamental to prevent machine-generated results from disagreeing with the instructor's intention.

\subsection{Natural Language: Multifaceted Challenge}

Using natural language to instruct machines may offer a false sense of ease, leading many to undervalue the need for crucial skills for efficient machine instruction. Unlike programming languages, designed with strict syntax and semantics, natural languages are inherently complex and ambiguous, filled with idioms and cultural nuances, making full comprehension by machines challenging \citep{pratt2010computational}. 

Interacting with machines through natural language has become significantly more accessible from a user's perspective. However, for those aiming to work as prompt engineers and derive value from pre-trained models through instructions, the complexity of natural language presents unique challenges not found in programming languages. This complexity requires a deep understanding of linguistic nuances, far beyond the simplicity of user-level interactions, emphasizing the need for specialized skills to navigate and leverage natural language effectively in the realm of prompt engineering.

The richness of vocabulary with its synonyms, antonyms, and polysemic terms significantly influences text interpretation. Complex grammar adds another layer of difficulty, requiring accurate comprehension of syntactic and semantic relationships. Ambiguity, both lexical and structural, introduces significant interpretive complexity, often necessitating context and linguistic cues analysis \citep{savitch2012formal}. Figures of speech like metaphors and ironies, intrinsic to literary expression, enrich texts but complicate direct interpretation, requiring a deep understanding of textual and cultural contexts. Historical, cultural, and social references in texts demand familiarity with the language, culture, and history \citep{harras2000concepts}. 

Text interpretation is complex, underscoring the importance of not oversimplifying these complexities when instructing machines in natural language and considering the nuances of language interpretation before attributing errors to the machine.

Thus, if mastering a programming language with rigid syntax requires skill, navigating the complexities of natural language, with its inherent ambiguities, demands expertise.

\subsubsection{Important concepts in linguistics}

To improve the clarity of prompts, it is important to have a good understanding of the language. This involves tailoring the prompt's language to get precise answers, accurately interpreting the model's responses in complex scenarios, and recognizing when a response might have misinterpreted your intended meaning due to subtle language details. 

In linguistics, key concepts such as Morphology, Syntax, Semantics, and Pragmatics \citep{rosa2013introduction}, are crucial when instructing a machine, being:

\textbf{Morphology:} The study of the structure and formation of words, including the analysis of morphemes (the smallest grammatical unit of meaning or grammatical function) \citep{haspelmath2013understanding}. Although LLMs are trained on vast amounts of text and therefore have a good understanding of word morphology, a clear understanding of this concept can help you formulate prompts that consider morphological variations. This is useful for creating prompts that are specific enough to obtain the desired responses, and recognizing the relationship between different forms of a word.

\textbf{Syntax:} The study of sentence structure and the rules that govern sentence formation in a language \citep{tallerman2019understanding}. LLMs can understand and generate text based on syntactic rules due to their training in large text corpora. However, understanding syntax can help you structure prompts and interpret responses more effectively, especially for complex tasks that require deep understanding or structured text generation.

\textbf{Semantics:} The study of meaning in languages, covering both the meaning of individual words and the meaning generated by the context of sentences \citep{lobner2014understanding}. This concept is crucial for interacting with LLMs, as it allows you to create prompts that align with the desired meaning. Understanding semantics can help avoid ambiguities and improve the accuracy of model responses, especially when dealing with polysemous words or abstract concepts.

\textbf{Pragmatics:} The study of how context influences the interpretation of meaning, focusing on how people use language in specific communicative contexts and how the principles of language use contribute to effective communication \citep{leech2016principles}. Although LLMs may not have the capacity to fully understand the context or implicit intentions as humans do, knowing pragmatics is useful for shaping prompts in a way that anticipates and guides the model's interpretation. This is especially important for tasks involving nuances of communication.

\textbf{Structural and lexical ambiguity:} Structural and lexical ambiguity is important, particularly in the study of semantics and syntax, referring to different types of ambiguities found in language \citep{sowa1993lexical}.

\textbf{Lexical Ambiguity:} Lexical ambiguity occurs when a word has more than one possible meaning. This can lead to confusion in interpreting sentences where polysemous words (words with multiple meanings) are used without clear context specifying which meaning is intended. For example, the word "bank" can refer to a financial institution or the side of a river; without additional context, one cannot determine which meaning is correct \citep{rodd2018lexical}.

\textbf{Structural Ambiguity:} Structural ambiguity, also known as syntactic ambiguity, occurs when a sentence can be structured in more than one way, leading to more than one possible interpretation. This is usually due to how words are arranged in the sentence, which can generate different interpretations of which words or phrases are related to each other \citep{bustam2016analysis}. An example is the sentence  \textit{"I saw the man with the telescope"}, which can be interpreted in two ways: the person observing uses a telescope to see the man, or the person sees a man who is holding a telescope. 

Both types of ambiguity are central to the study of language, as they illustrate how human communication can be complex and subject to misunderstandings. They are also relevant to fields like natural language processing, where disambiguating the meaning of words and sentences is crucial for correct language understanding and generation by computers.

\begin{figure}[htb!]
\centering
\begin{subfigure}[b]{0.3\textwidth}
\centering
\includegraphics[width=\textwidth]{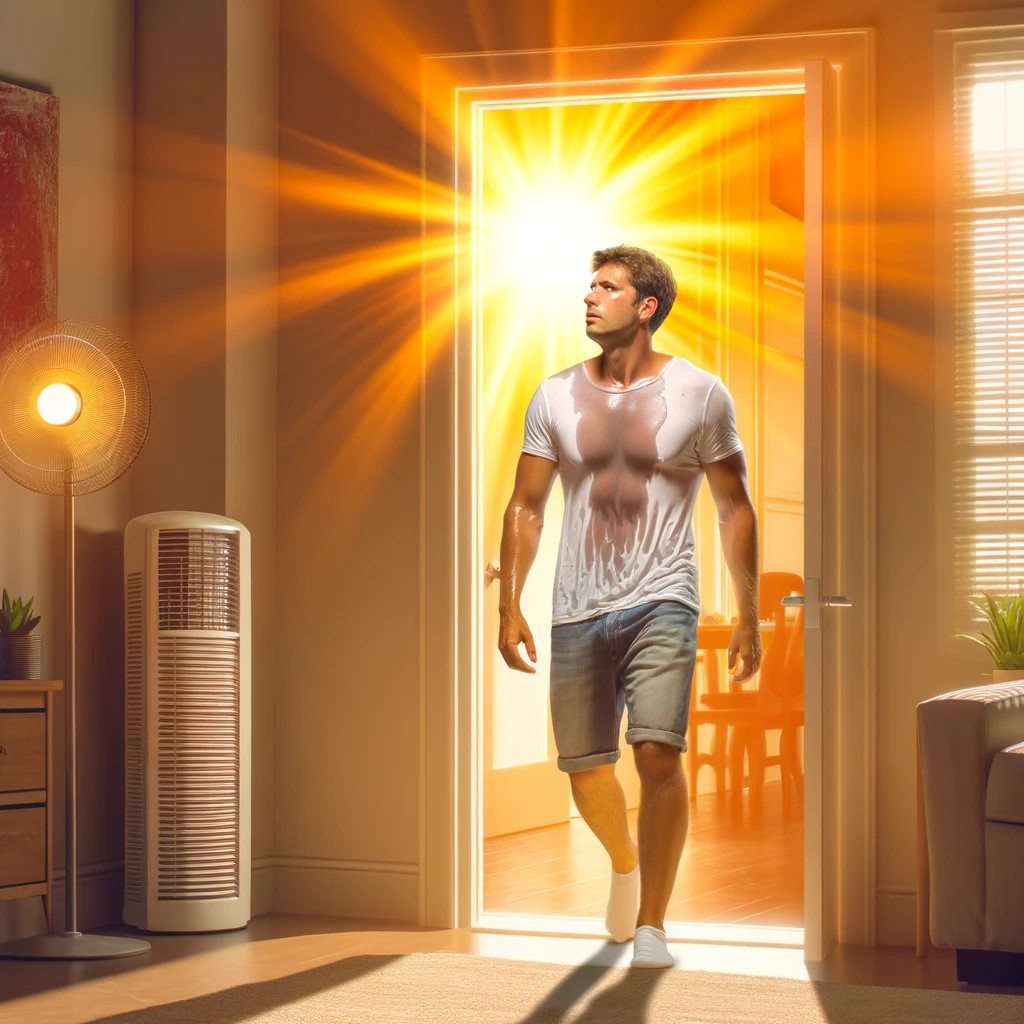}
\caption{Interpretation: Paulo leaving a room that is physically hot, with an emphasis on his personal discomfort from the heat.}
\label{fig:sub1}
\end{subfigure}
\hfill
\begin{subfigure}[b]{0.3\textwidth}
\centering
\includegraphics[width=\textwidth]{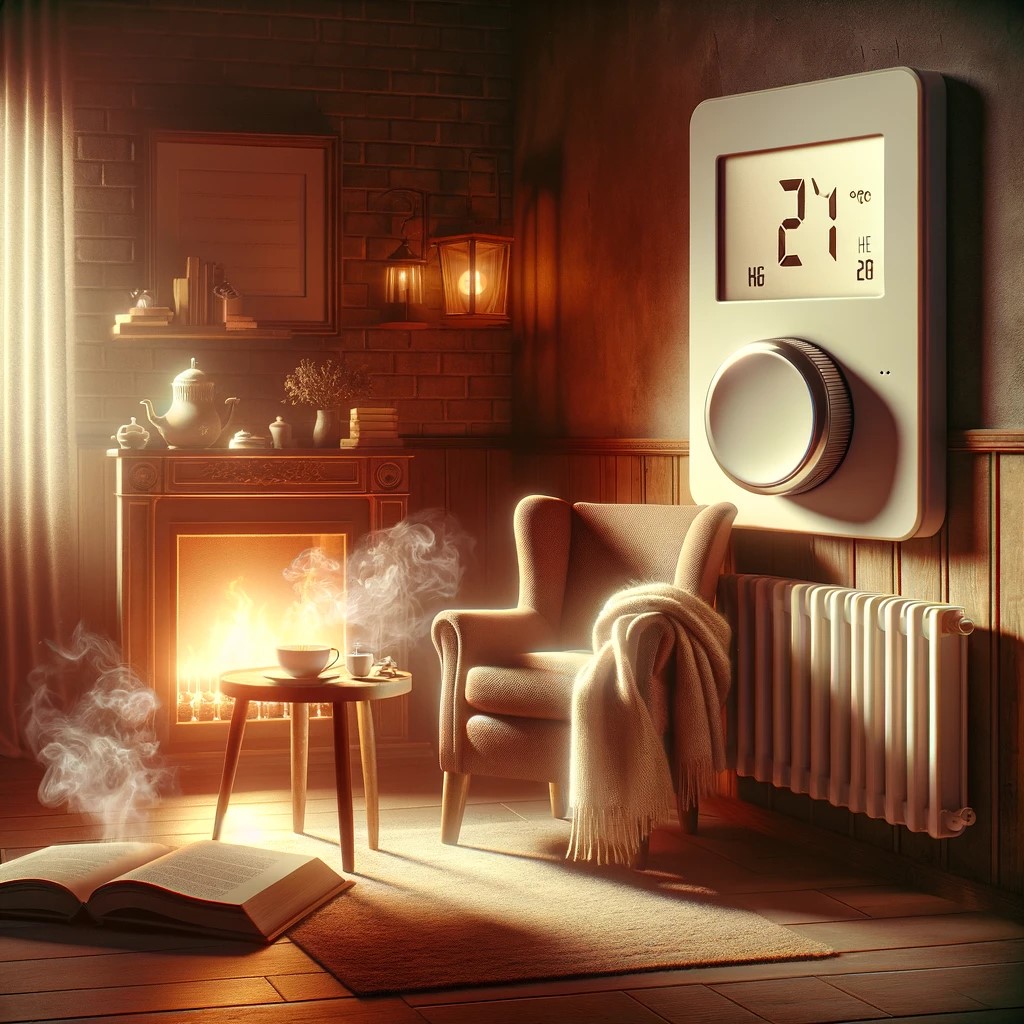}
\caption{Interpretation: An empty room left hot, focusing on the warm, cozy atmosphere created by a high thermostat setting or a heating element.}
\label{fig:sub2}
\end{subfigure}
\hfill
\begin{subfigure}[b]{0.3\textwidth}
\centering
\includegraphics[width=\textwidth]{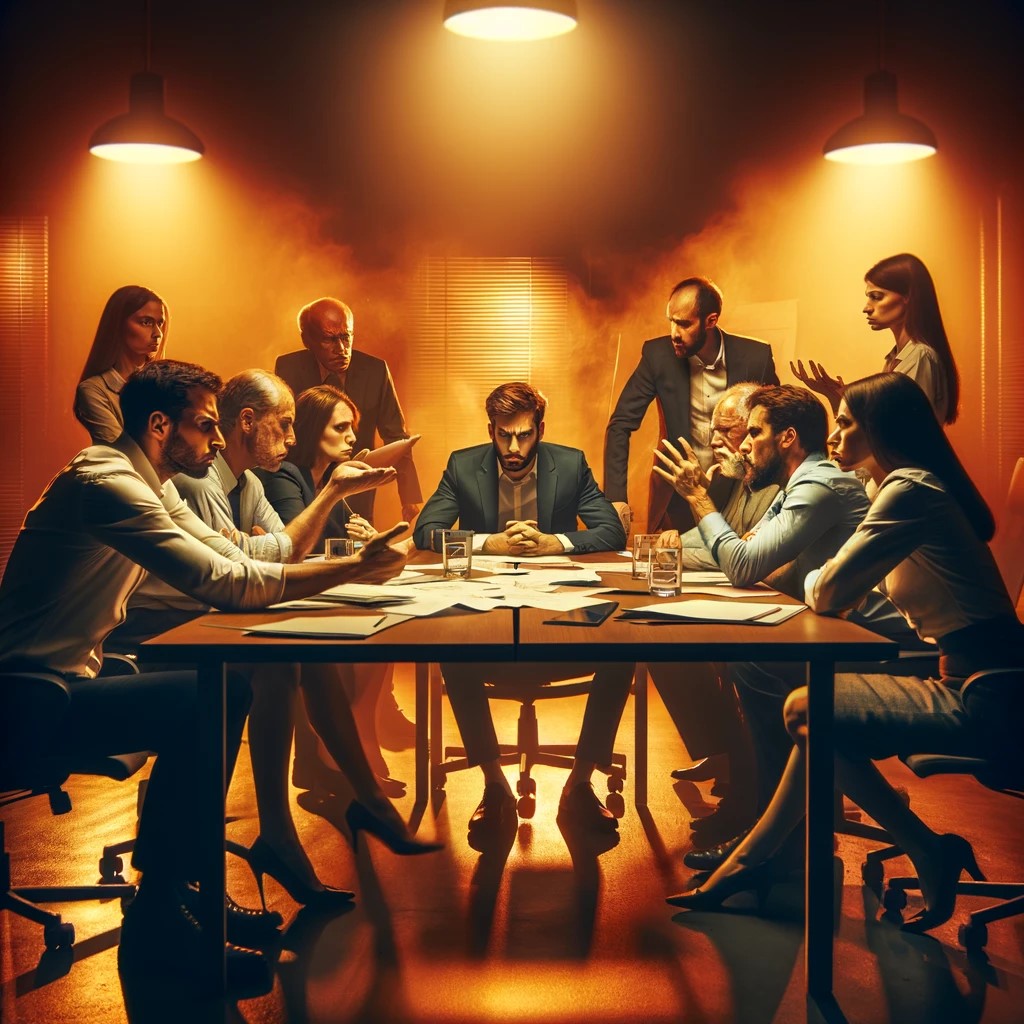}
\caption{Interpretation: A tense and emotionally charged situation in a room, depicted through the body language, suggesting a heated discussion}
\label{fig:sub3}
\end{subfigure}
\caption{"\textbf{Paulo left the room hot"}: The ambiguity primarily arises because "left" can be interpreted both as the act of leaving a place and the state in which something was left. Additionally, "hot" can describe either the physical temperature of an object or environment or an emotional or situational state.}
\label{fig:overall}
\end{figure}

\begin{figure}[htb!]
\centering
\begin{subfigure}[b]{0.4\textwidth}
\centering
\includegraphics[width=\textwidth]{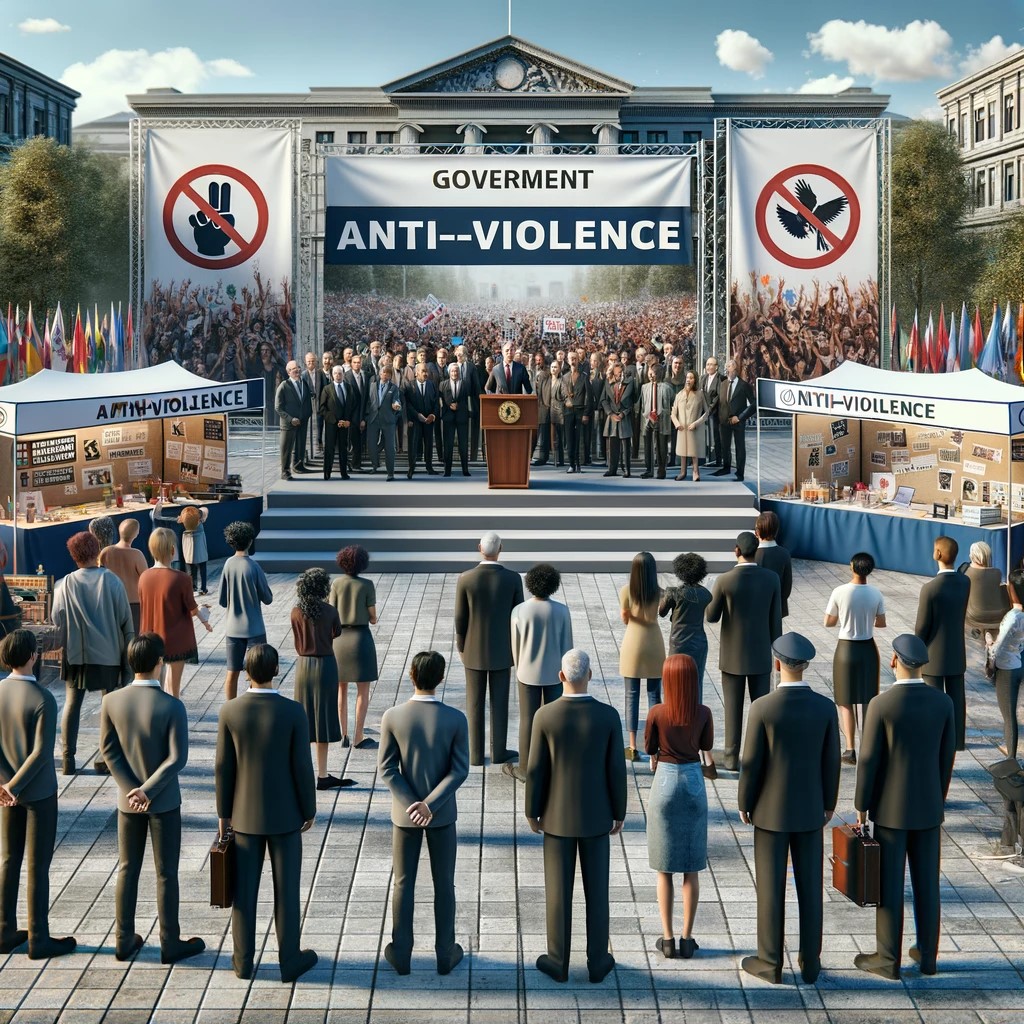}
\caption{Government-Led Campaign Against Violence:
The government has initiated a campaign aimed at combating violence within society. The campaign's goal, for the government to act as the promoter of initiatives to reduce or prevent violence.}
\label{fig:sub1}
\end{subfigure}
\hfill
\begin{subfigure}[b]{0.4\textwidth}
\centering
\includegraphics[width=\textwidth]{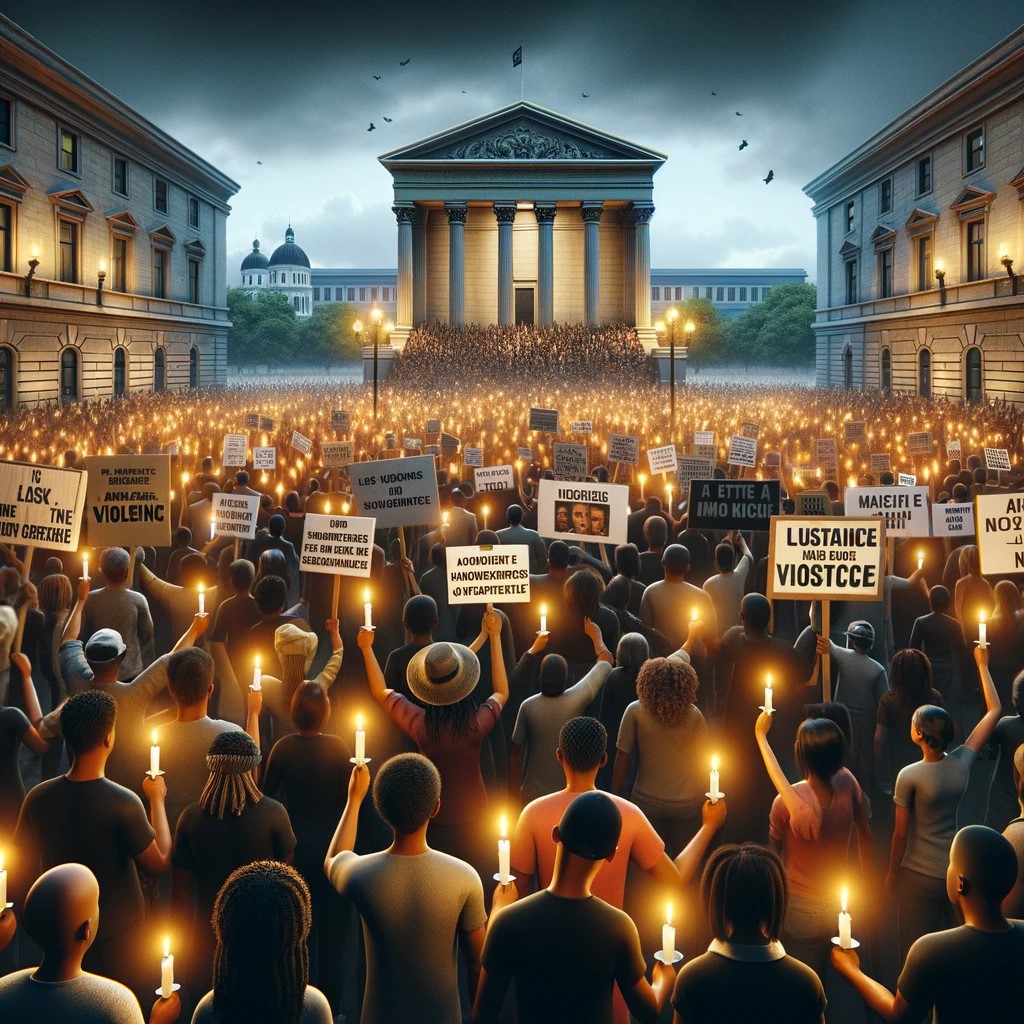}
\caption{Campaign Against Government-Perpetrated Violence: The campaign as aiming to combat violence perpetrated or endorsed by the government itself. This could include police brutality, suppression of protests, or other forms of power abuse.}
\label{fig:sub2}
\end{subfigure}
\caption{"\textbf{The campaign on government violence has initiated"}: The ambiguity here lies in the phrase "on government violence," which does not clarify whether the campaign is aimed at fighting violence from the government or is a campaign conducted by the government against violence. This formulation maintains an element of ambiguity regarding the government's role in the campaign.}
\label{fig:overall2}
\end{figure}

 To illustrate the complexity of interpreting natural language, let us analyze two sentences: "Paulo left the room hot" (see Figure \ref{fig:overall}) and "The campaign on government violence has initiated" (see Figure \ref{fig:overall2}).

Note that, the issue of interpretation complexity can lean towards any side, depending on the reader's perspective. This openness to inference may lead to interpretations that may not align with the writer's intended meaning. These examples highlight the inherent ambiguity in language.

The potential for linguistic ambiguity can skyrocket when leveraging models not trained in one's native language to extract business value \citep{chen2023monolingual}. This increase becomes particularly pronounced in instances where sentences are crafted in a language in which the author is not fluent. The cultural nuances, idiomatic expressions, and syntactic structures that might appear clear-cut to native speakers can transform into obstacles for non-natives \citep{nursidi2019monolingual}.

This linguistic gap significantly heightens the risk of misinterpretation. Understanding the complexities of language and its inherent potential for ambiguity is paramount when directing language models. Analogous to human interaction, language models necessitate explicit and accurate instructions to produce text that is both correct and contextually relevant. 

For those instructing the machine, acquiring an understanding of the grammar, syntax, semantics, and cultural nuances of the target language is important. Moreover, it is vital to recognize the limitations and potential stumbling blocks associated with language models due to training in another language, particularly when they are applied to business landscapes. Language models may falter when encountering idiomatic expressions, cultural allusions, and linguistic nuances that diverge from those present in their training datasets. 

It is crucial to carefully evaluate the scenario in which one chooses to use a model not trained in the native language, as it may lack the capability to interpret basic relationships in another language.

Often, companies employ an automatic translation layer to overcome this barrier, but this approach can be extremely inefficient in business scenarios. If instructing the machine in the native language introduces the inherent complexities of natural language, adding a layer of abstraction through automatic translation can increase the risks of receiving responses that do not align with expectations.

Although linguistic models offer significant opportunities for business applications and communication, the decision to use a model not trained in the target language should not be taken lightly. The complexities introduced by language barriers, especially when compounded by automatic translation, can lead to misinterpretations and substantial inefficiencies. These challenges underscore the importance of choosing the right tools for the task at hand, preferably opting for models that have been trained in the native language of the application whenever possible.

If the use of a non-native model is unavoidable, it becomes essential to implement rigorous testing and validation processes to mitigate potential misunderstandings and ensure that the model's outputs closely align with the intended meanings and cultural nuances. Only through careful consideration and strategic planning can businesses leverage the full potential of linguistic models while minimizing the risks associated with interlinguistic misunderstandings.

The misinterpretation of machine-generated responses due to inadequate prompt engineering should not be construed as "hallucinations" on the part of the machine. Such a perspective overlooks the nuanced complexities inherent in written language, attributing undue error to the machine rather than acknowledging the prompt engineer's failure to provide clear and comprehensive directions. 

This scenario mirrors the principles observed in programming languages, where a failure to adhere to syntactic rules results in code errors, reflecting not on the inadequacies of the programming language but on the programmer's lack of precision.

In the realm of natural language, a similar paradigm applies; the transition from the rigid syntax of programming languages to the more abstract syntax of natural language does not diminish the importance of precise instruction. For instance, consider the Portuguese prompt: \textit{"Especifique um \textcolor{red}{Câmbio} para uma empresa automotiva."}.

This prompt, devoid of clear context, leaves uncertain whether "Câmbio" refers to "foreign exchange" or "vehicle transmission." Such ambiguity exemplifies the instructor's failure to specify the intended meaning, leading to potential misinterpretation by the machine.

\begin{figure}[htb!]
\centering
\includegraphics{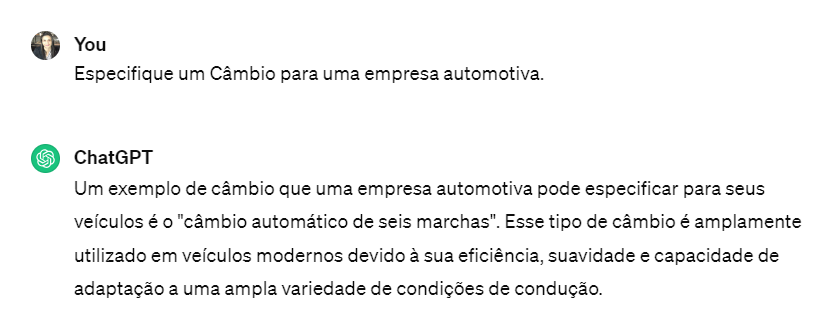}
\caption{Interpretation of "Cambio" as "vehicle transmission" by the machine, showcasing the importance of context in guiding machine interpretation.} \label{cambiofig}
\end{figure}

As illustrated in Figure \ref{cambiofig}, the machine's interpretation of "Câmbio" as "vehicle transmission" underscores the role of linguistic context and pragmatic semantics in guiding machine understanding.

Using linguistic knowledge, one can conclude that attributing "transmission" is the most probable inference for several reasons:

Linguistic Context: The word "câmbio" in Portuguese can refer to both currency exchange and vehicle transmission. In the context of an automotive company, it is more common to refer to transmission, as it is an essential part of the vehicles they manufacture.

Pragmatic Semantics: Considering the overall context of the request (an automotive company), the model may have inferred that "câmbio" more likely refers to the mechanical part of the vehicle, i.e., transmission, rather than financial currency exchange operations.

This interpretation, predicated on the statistical likelihood of word usage within the provided context, exemplifies the machine's reliance on linguistic knowledge and contextual cues for semantic inference.

In other instances, consider the English prompt, where the instructor's intentions are not explicit in a sports context:
\textit{"How does a \textcolor{red}{bat} navigate?"}.

This question could lead to errors because "bat" can refer to both the animal and a piece of sports equipment. Without specifying the context, the LLM might explain the sports equipment's design and use instead of the animal's echolocation abilities, demonstrating how crucial precise wording and context are to ensure accurate interpretation by an LLM.

\begin{figure}[htb!]
\centering
\includegraphics{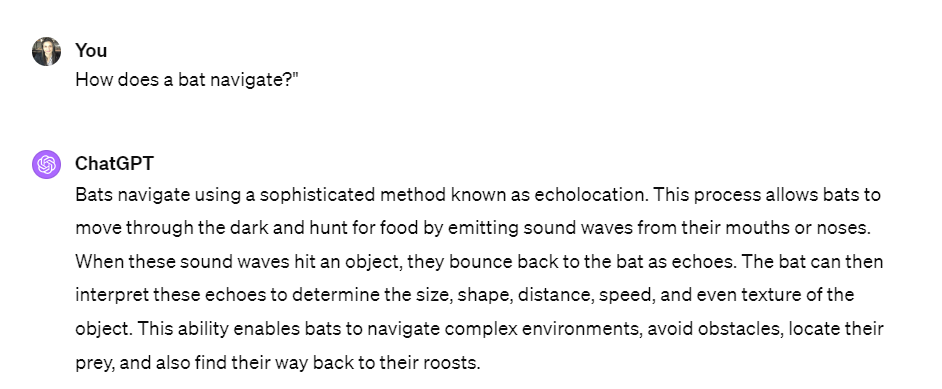}
\caption{Interpretation of "bat" as "animal" by the machine, showcasing the importance of context in guiding machine interpretation.} \label{cambiofig}
\end{figure}

These examples underscore the critical need for clarity in instruction to prevent misalignments between the instructor's intentions and the machine's responses.

It is imperative to acknowledge that machine interpretations are based on explicit instructions and the probabilistic nature of language understanding. 

Thus, many discrepancies can be attributed to errors in direction, rather than flaws in the machine's operational logic. The language models, in the examples "Câmbio" and "bat" context, probability may have encountered numerous instances of "bat" in animal contexts and "Câmbio" in automotive contexts during its training, suggesting a tendency towards these interpretations. While such assumptions about the training data are speculative, since it is not possible to know the exact nature of the data used to train these models, they highlight the importance of considering the machine's probabilistic inference mechanisms when providing instructions. 

Relying on the machine to make an inference on the most probable context is akin to relying on the luck of its probabilities being aligned with the instructor's intention. 

The advice \textit{"it is important to contextualize"} translates to the importance of paying attention to linguistic concepts and assuming the responsibility that the "hallucinations" attributed to the machine are often failures of the instructor to provide unambiguous guidance.

Understanding linguistic concepts, regardless of language, reinforces the importance of accurate communication in human-machine interactions.

\section{Influencing the output of pre-trained generalist language models}

An efficient way to specialize a pre-trained model for a specific task is through fine-tuning. In the context of machine learning, particularly with reference to language models like GPT (Generative Pre-trained Transformer), fine-tuning refers to the process of taking a pre-trained model and further training it on a smaller, specific dataset to adapt it for a particular task or domain. This method leverages the general knowledge the model has acquired during its initial extensive training on a large dataset, enabling it to perform well on tasks not covered during that training \citep{lankford2023adaptmllm}.

While fine-tuning is a powerful practice for enhancing model performance on specific applications, it comes with several complexities and challenges:

Data Availability - Quality and Quantity: For effective fine-tuning, it is necessary to have a high-quality and relevant dataset for the specific task. The amount of data is also crucial; insufficient data can lead to overfitting, while too much data can make the process costly and time-consuming. The fine-tuning data must be representative of the specific task or domain to ensure that the adjusted model generalizes well to new examples within that context \citep{wang2024survey}

Overfitting - Balance between Generalization and Specificity: Fine-tuning can cause the model to become overly specialized on the training data, losing its ability to generalize to new data. This is particularly problematic in small or highly idiosyncratic datasets,  meaning datasets that have peculiar characteristics \citep{xue2023repeat}.

Computational Complexity - Required Resources: Fine-tuning can be computationally intensive, requiring high-performance hardware (such as GPUs or TPUs) and significant time, especially for very large models \citep{xu2024survey}.

Choice of Hyperparameters - Learning Rate and Regularization: Adjusting the learning rate and regularization methods (such as dropout) is crucial to avoid overfitting and underfitting. Finding the right set of hyperparameters can be a time-consuming process of trial and error \citep{wang2023cost}.

Knowledge Degradation: A common problem in fine-tuning is catastrophic forgetting, where the model loses its ability to perform the tasks for which it was originally trained by focusing too much on the new training data \citep{zhai2023investigating}. 

Fine-tuning has emerged as a crucial technique in the deployment of machine-learning models for specialized tasks. It enables the utilization of the vast capabilities of large, general models for specific \citep{lin2024data}. 

However, not all users and companies can navigate the complexities and challenges associated with fine-tuning due to constraints such as limited technical knowledge, inadequate infrastructure, and insufficient data.

\subsection{Customizing Attention Mechanisms}

A strategy available to all is to utilize the capability of steering the focus of these models by applying two critical skills: an understanding of information processing and a mastery of natural language linguistics.

\begin{figure}[htb!]
\centering
\includegraphics[width=0.9\textwidth]{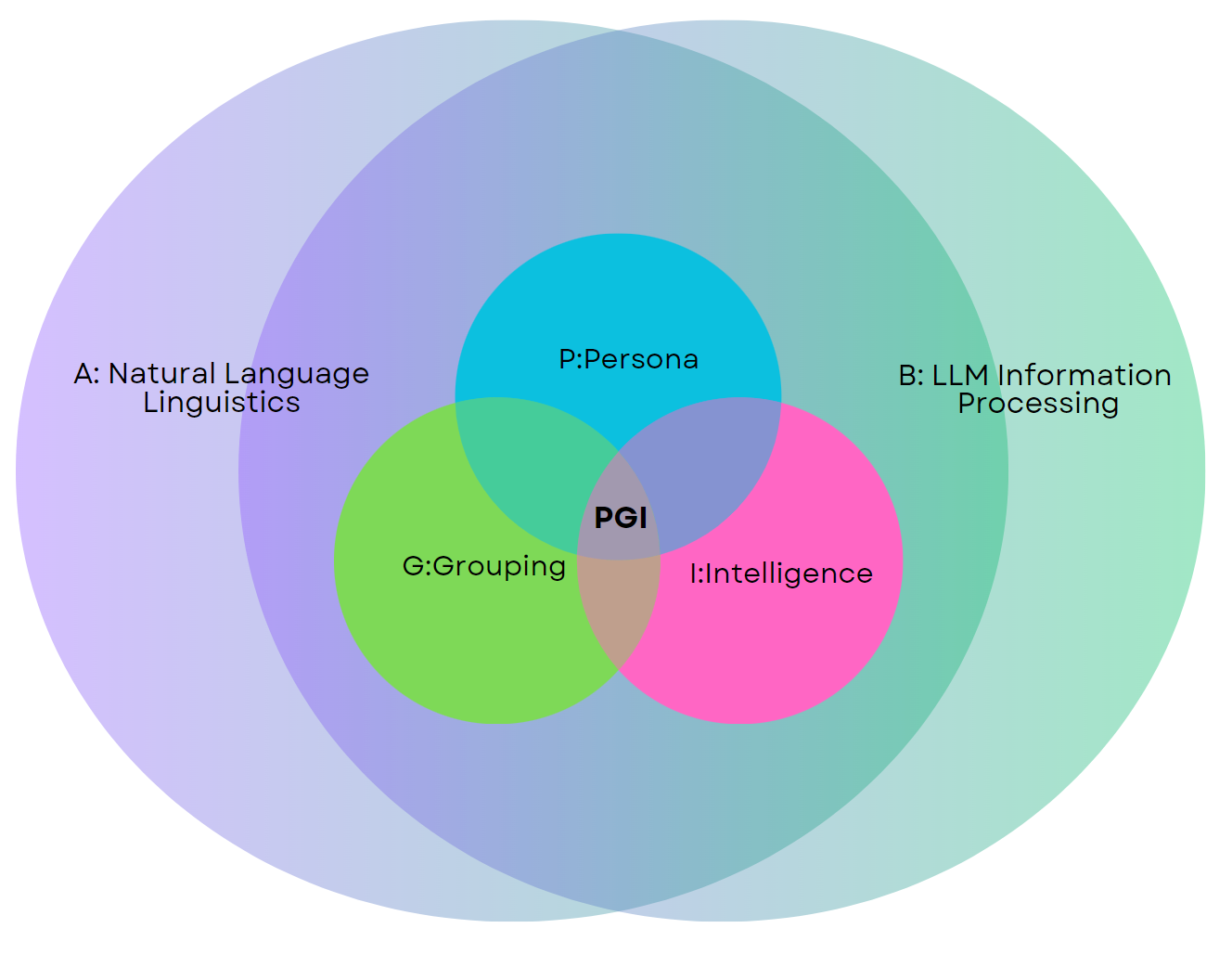}
\caption{PGI - Triple Bottom Line in Attentional Guidance: Bridging the Two Pillars of Stability through Natural Language Linguistics and LLM Information Processing} \label{pci2}
\end{figure}

The Persona-Grouping-Intelligence (PGI) framework was designed to solve a specific pain point in business: keeping the attention of LLMs focused on specific business issues \citep{2023arXiv230813317I}. The framework leverages the sophisticated text comprehension and generation capabilities of LLMs, directing these abilities to effectively address defined business challenges. PGI does not offer prompt tips; it fosters a logic for creating prompts based on a solid foundation of Natural Language Linguistics and an understanding of LLM Information Processing. Within this framework, PGI emphasizes the importance of Persona-Grouping-Intelligence, endorsing an approach that employs deep linguistic insights and LLM information processing techniques to improve LLM functionality and applicability, as illustrated in figure \ref{pci2}.

Subsequent sections will delve deeper into this paradigm shift, focusing on the integration of these new skills and how each interconnection within the PGI framework is designed to influence the outcomes of LLM models. This philosophy and the best practices for employing each intersection within the PGI structure will be outlined.

\subsubsection{Intersecting Domains: Natural Language Linguistics, LLM Information Processing, and \textsc{PERSONA} Customization}

Within the PGI framework, "Persona" is defined as a technical construct that directs the model's focus toward a specific context \citep{2023arXiv230813317I}.

These personas are crafted to refine the model's output, ensuring precise alignment with the preferences and requirements of individual users, tasks, or domains. A well-conceived persona serves as a knowledge filter, markedly improving how the model interacts with users. However, the development of these personas often forgoes a solid foundation in linguistic principles and a comprehensive understanding of the model's language processing capabilities.

Constructing a persona should be firmly based on linguistic theories and the operational dynamics of the model. Essentially, a persona filters the model's extensive knowledge base, favoring words and concepts associated with it during the generation process, which results in more contextually relevant responses.

LLMs possess a broad spectrum of knowledge spanning various fields, due to their training on extensive datasets \citep{zhuang2024toolqa}. When devising a persona, the aim is to influence the model's attention during generation, guiding it to understand semantics—the meanings of words and phrases within a specific context—through the perspective of the persona.

This approach not only shapes semantic interpretation but also pragmatics, involving the understanding of language as used in context. A thoughtfully designed persona leads to nuanced interpretations of requests, personalized dialogues, and responses that understand the implied meanings.

A persona should transcend a mere list of attributes, embodying a comprehensive behavioral and contextual framework that enables the LLM model to interact in a manner reflecting the persona's characteristics. This alignment enhances the model's semantic and pragmatic interpretations, closely matching the intended user experience. Explicitly defining a persona's nationality or language within a multilingual model is critical to ensure consistent interaction in the chosen language, avoiding confusing language switches \citep{2023arXiv230813317I}.

\begin{figure}[htb!]
\centering
\includegraphics[width=0.89\textwidth]{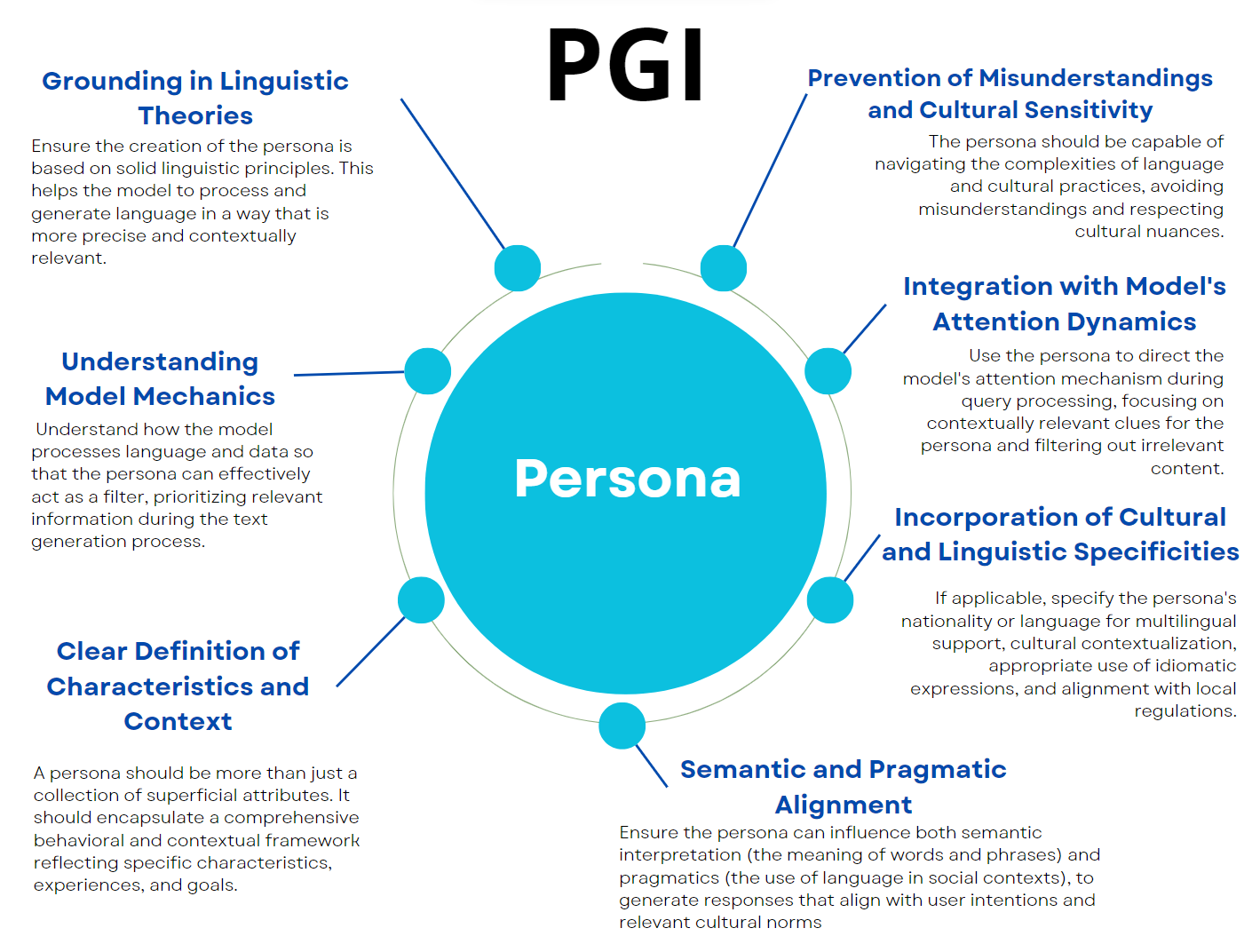}
\caption{Intersection: Natural Language Linguistics X LLM Information Processing. The best practice for defining a persona in PGI involves a detailed process that incorporates both linguistic knowledge and an understanding of the model's operational mechanics. Key steps for defining a persona effectively.} \label{PersonaPGI}
\end{figure}

To illustrate the theory, let us consider "FinAI," a persona that embodies a knowledgeable Brazilian banking professional, crafted to provide users with tailored financial advice and services. FinAI transcends mere collections of banking keywords; it is a character infused with professionalism, a formal tone, regulatory knowledge, and expertise in financial products. This persona's Brazilian nationality adds a layer of cultural and regulatory specificity, ensuring that financial advice is not only technically sound but also culturally relevant and compliant with local laws and practices.

FinAI's persona significantly influences the model's word and concept representations, prioritizing banking terms like "investment" in the semantic space. This specificity guarantees that words are interpreted within the appropriate banking context, thus enhancing their relevance. The inclusion of Brazilian cultural and regulatory nuances ensures that the model's responses are aligned with local expectations and norms, further personalizing the user experience.

When processing queries, the model's attention mechanism focuses on cues aligned with FinAI's expertise, incorporating the Brazilian financial context. This persona-based interpretation ensures that queries are analyzed through a financial lens, taking into account potential goals and the Brazilian regulatory environment. Consequently, a term like "secure" might be associated with financial security within Brazil's unique context, directing the model to provide responses that resonate with FinAI's persona.

Incorporating linguistic principles into persona development is crucial for preventing misinterpretations and enhancing the model's output accuracy. A well-defined persona, such as FinAI, ensures contextual precision by associating terms like "withdrawal" specifically with financial transactions within the Brazilian banking sector. It aids in the disambiguation of complex terms, facilitates pragmatic appropriateness by adhering to the persona's sector, influences the generation of contextually relevant responses, and maintains a voice and responses reflective of the persona's characteristics.

Specifying FinAI's Brazilian nationality within the persona is paramount for ensuring the multilingual model interacts consistently in Portuguese, avoiding confusing language switches, and grounding financial advice in the Brazilian context. This approach underscores the importance of cultural sensitivity, recognizing the specific terminologies, practices, and regulatory frameworks unique to Brazil.

The best practice for defining a persona, in PGI, involves careful and informed development, ranging from theoretical foundations to practical application as shown in Figure \ref{PersonaPGI}

In summary, the strategic development of AI personas, exemplified by FinAI, is not merely a sophistication—it is essential for operational accuracy and the reliable functioning of AI in domain-specific applications like banking. By grounding a persona in linguistic concepts and specifying its nationality in the multilingual model, it is better equipped to understand and process language accurately, respecting the complexity of language communication within a given cultural and regulatory context. This method ensures that LLMs like GPT can offer personalized, relevant, and culturally attuned interactions, enhancing the overall user experience.

\subsubsection{Intersecting Domains: Natural Language Linguistics, LLM Information Processing, and Intelligence}

Within the PGI framework, "Intelligence" is defined as the strategic extraction of business insights. This ensures that the generated data by the model is not only coherent but also precisely aligned with the specific expectations of the business environment \citep{2023arXiv230813317I}.

The process of human thought is deeply rooted in context, often swayed by emotional states or nuanced cues that play a crucial role in shaping interpretation and decision-making processes \citep{sternberg1988psychology}. This complexity presents a challenge for LLM models, which, despite their ability to process linguistic signals to a certain degree, struggle to fully understand the breadth of human emotions or the subtleties of contextual backgrounds \citep{zhong2019knowledge}. Consequently, their responses, while technically correct, may lack the necessary empathy, cultural sensitivity, or appropriateness to fully meet human expectations.

In the rapidly evolving landscape of artificial intelligence, there is a growing demand for prompt engineers who possess a blend of technical skills and the ability to adapt behavior to leverage pre-trained models effectively. However, many companies have not realized that extracting value from such models requires new skills. This new breed of professionals must possess a skill set that extends beyond traditional data science expertise, encompassing competencies in natural language linguistics and an understanding of how LLMs process information. 

For data scientists, a crucial shift in mindset is necessary, recognizing that crafting prompts for LLMs demands a set of skills markedly different from those used in traditional model development. Data scientists typically work with a level of separation from the business sectors, focusing on collecting data, training models, validating results with new datasets, and ultimately, delivering the final model to the business stakeholders. However, this approach, when applied in isolation from the business context to guide an LLM, can be notably ineffective. The reason is that data scientists may lack the necessary business acumen and understanding of the linguistic nuances essential for leveraging the full potential of LLMs. This gap can hinder their ability to fully utilize these advanced models to deliver results that meet or exceed business expectations.

Essential skills for an efficient prompt engineer include strong communication and collaboration abilities, as collaboration with business teams is paramount. The capability to articulate technical needs in a language understandable to business teams, and conversely, to translate business thinking into effective prompts, is crucial. Critical thinking is equally important, involving maturity to include taking responsibility for outputs that do not align with business expectations due to misdirected focus, as well as critically assessing potential biases and inaccuracies.

\begin{figure}[htb!]
\centering
\includegraphics[width=0.9\textwidth]{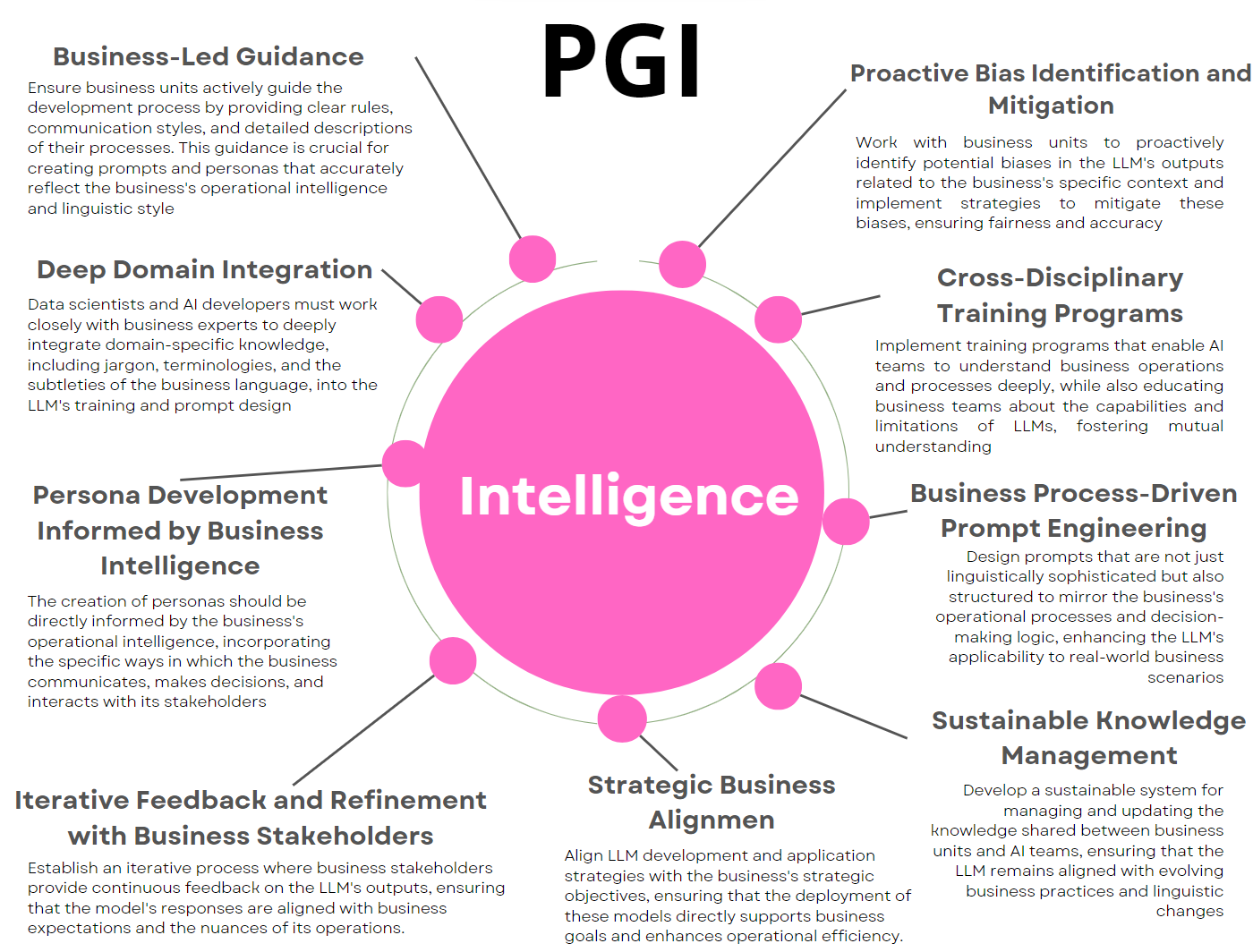}
\caption{Intersection: Natural Language Linguistics X LLM Information Processing. The best practice for defining Intelligence in PGI involves the collaboration between data scientists and business units to become more effective, ensuring that the development and application of LLMs are deeply rooted in the intelligence and linguistic specifics of the business domain. This approach not only enhances the technical performance of the models but also ensures their outputs are highly relevant and valuable to the business operations they are designed to support.} \label{IntelligencePGI}
\end{figure}

To truly address real-world business problems, there is a need for a deep understanding that crafting effective prompts requires leveraging linguistic concepts. This entails recognizing that to solve a business problem, without extracting intelligence and terms reflective of business experts' thinking, achieving desirable outcomes is unlikely. This comprehensive skill set is crucial for prompt engineers to harness the full potential of AI in business challenges.

In the context of PGI, best practices for intelligence extraction entail a meticulous and informed approach, extending from theoretical underpinnings to tangible implementation, as illustrated in Figure \ref{IntelligencePGI}.

Another important mindset shift is that empowering business teams to effectively utilize the LLMs marks a significant shift toward a more integrated and collaborative approach to leveraging artificial intelligence within organizations. This initiative involves equipping non-technical business professionals with the knowledge and tools necessary to create prompts that extract valuable insights and solutions from LLMs, aligning directly with their strategic objectives and challenges.

Organizations must recognize that prioritizing education and training offers the most direct path to achieving desired outcomes. Blind experimentation not only leads to wasted time and resources but also fails to equip teams with the necessary skills to extract value. The most efficient route is through the pursuit of knowledge, and encourage prompt engineering by business units.  This approach ensures that teams are not only well-informed but also capable of leveraging LLM technology to drive innovation.

By equipping business teams with the skills to leverage LLMs effectively, organizations can unlock new levels of efficiency, innovation, and competitive advantage. This democratization of AI tools fosters a culture of continuous improvement and adaptability, essential in the rapidly evolving business landscape.

Ultimately, blending business knowledge with AI technology not only boosts operational efficiency and drives innovation but also unlocks previously unattainable capabilities. Such strategic integration of linguistic processing and LLMs equips businesses to incorporate AI as a fundamental element of their strategic toolkit, demonstrating the significant impact of this convergence.

\subsubsection{Intersection: Natural Language Linguistics X LLM Information Processing X GROUPING}

Within the PGI framework, "Grouping" is identified as the contextual linguistic clustering that surrounds the model's operation. This aspect involves a deep understanding of language nuances, contextual implications, and the interrelations between words and the business problem \citep{2023arXiv230813317I}.

Transformer models represent a paradigm shift in the way machines process sequential data and chain thoughts \citep{vaswani2017attention}. This advancement underscores the transformative potential of Transformer models in pushing the boundaries of artificial intelligence and its applications in understanding and generating human language.

Despite these advancements, Transformer models still exhibit gaps in understanding human intelligence, meaning that to fully comprehend our intentions or what we expect from them, we need to pay attention to how we interact with these models to optimize their performance. 

The way humans work or think if presented with a language model with the same sequence of thoughts, might not be the most promising method. This is because machines have directed attention focuses and abstract many details that, while seemingly inconsequential to humans, can lead to responses that are logically correct but not aligned with our expectations.

Transformer models, despite their sophisticated attention mechanisms, are designed to prioritize information based on learned patterns from data \citep{vaswani2017attention}. This directed attention differs from human cognitive processes, where abstraction allows for a more nuanced filtering of information based on context, relevance, and even emotional cues. Humans can easily ignore irrelevant details or make leaps in logic based on shared cultural or contextual knowledge \citep{wang2010cognitive}, a subtlety that Transformer models may not always replicate accurately.

One of the challenges in interacting with Transformer models lies in the distinction between explicit and implicit knowledge. Humans often rely on implicit understanding or assumptions that are not directly stated but are understood within a context. 

For Transformer models to align more closely with human expectations, interactions must often be more explicit, detailing the context or intention that a human might deem unnecessary to articulate. This necessitates a thoughtful design in prompt engineering to guide the model toward the intended interpretation or response.

To address these gaps in thought processing, the "grouping" in PGI, emphasizes the need to present sequences of thoughts in a manner that simplifies comprehension for the machine. Specifically, in tasks that require multiple steps to be executed by a model LLM. These steps should be organized in a way that maintains the machine's focus because abrupt changes in context can confuse the model. It's crucial to remember that, despite appearing "intelligent," machines lack any awareness of the input text or the output they generate; they merely follow learned patterns and extract information based on probabilistic calculations. 

Machines excel in following linear and logical sequences but face difficulties with tasks that demand intuitive leaps or connections that a human might find obvious. When a prompt requires the machine to bridge gaps in logic not explicitly outlined in the input data, it may result in responses that are technically coherent but fail to capture the intended meaning or action. This challenge is partly due to the machine's inability to draw on broader contextual knowledge or shared human experiences that often inform our intuitive leaps.

LLMs determine relevance and focus based on the distribution of attention across the input sequence, guided by learned patterns of language. When the flow of information involves abrupt changes or requires shifting focus rapidly among unrelated concepts, the model's attention can become dispersed. This dispersion dilutes the model's ability to concentrate on relevant details. Attention dispersion can particularly impact tasks requiring sustained focus on a single narrative or logical thread.

Structuring information in a clear, logical sequence and minimizing unnecessary contextual shifts can help align the model's processing capabilities with the task's requirements. 

\begin{figure}[htb!]
\centering
\includegraphics[width=0.89\textwidth]{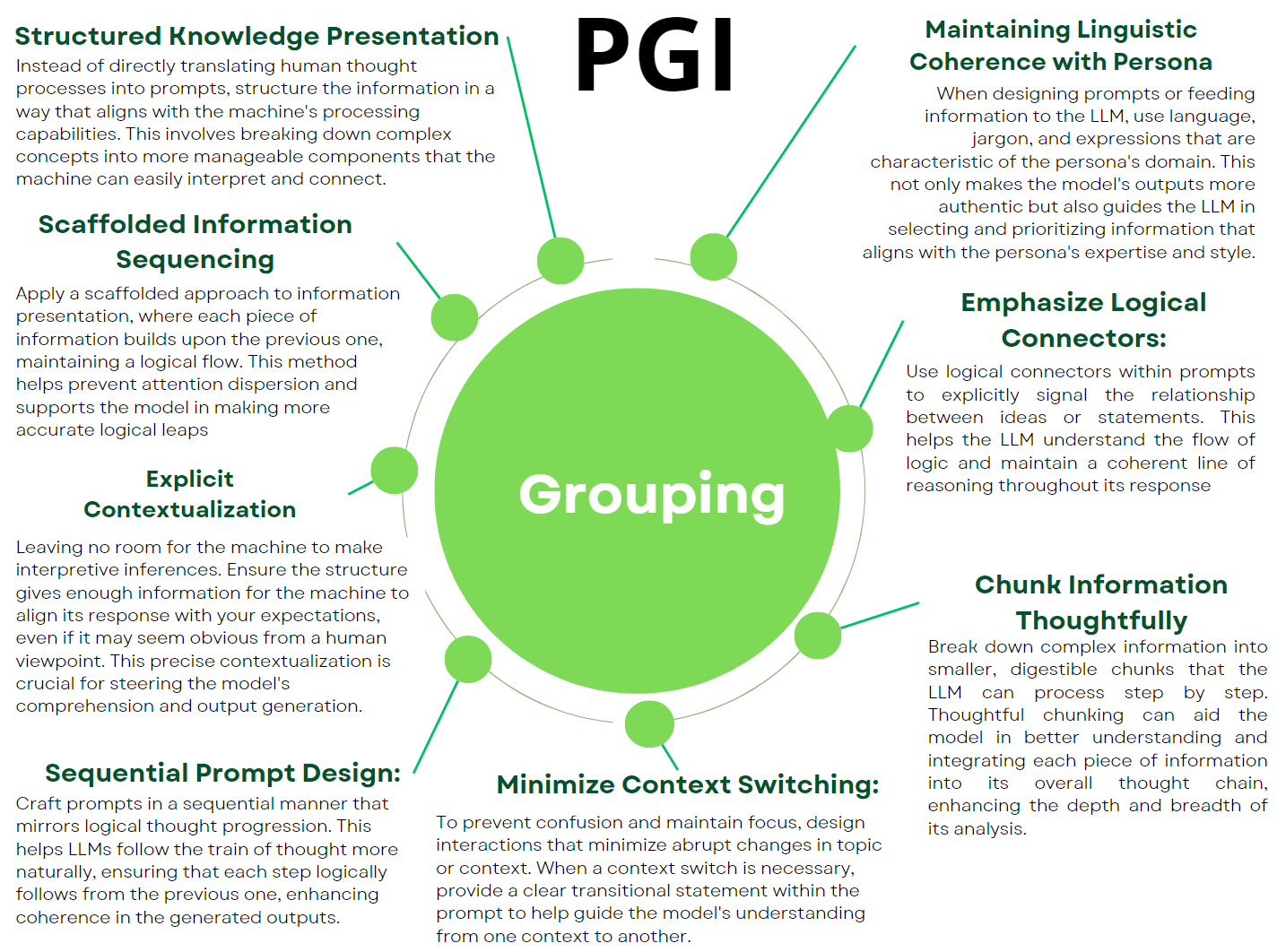}
\caption{Intersection: Natural Language Linguistics X LLM Information Processing. The best practice for defining Intelligence in PGI involves the process of extracting and implementing human intelligence in machine operations that can be optimized. This approach not only enhances the effectiveness of LLMs in performing complex tasks but also ensures that the outputs are more aligned with human expectations, bridging the gap between human cognitive processes and machine learning capabilities.} \label{Grouping}
\end{figure}

It is crucial to highlight that merely replicating the process of human intelligence extraction in the same manner is not a best practice. This approach overlooks the essential principle of leveraging the interconnectedness inherent in the PGI method, which lies at the intersection of linguistic knowledge and the machine's information-processing capabilities. The best practices, therefore, revolve around utilizing these interconnections effectively, as shown in figure \ref{Grouping}.

To bridge this understanding, the practice of grouping, within PGI, specifically, for tasks that demand a sequence of actions from an LLM, coherently organizing these actions is essential to maintain focus and avoid confusion \citep{2023arXiv230813317I}.

Large Language Models allocate attention based on the distribution of focus across the input sequence, relying on patterns of language they have learned  \citep{naveed2023comprehensive}. When confronted with rapid shifts in focus or unrelated concepts, the model's attention can become scattered,  being particularly detrimental to tasks that require a sustained focus on a coherent narrative or logical sequence.

For example, consider a task where an LLM is asked to analyze customer feedback across various cultural contexts to suggest improvements for a global product line. The model needs to navigate through diverse linguistic cues, emotional expressions, and cultural nuances to provide actionable insights. 

Prompt Example: \textit{"Read these customer reviews from the USA, Japan, France, and Brazil. Some people like the product's design but think it's too expensive. Others enjoy the functionality but not the color options. There are also comments about shipping times. By the way, did we mention the new feature launched last month? What do you think about implementing AI in our products? Also, consider the cultural differences in color preference and how the weather might affect product use in different regions."}

Note that: 

1: The prompt is unfocused, touching on multiple, loosely connected topics (product design, cost, functionality, color options, shipping times, new features, AI implementation, cultural differences, and weather impacts) without a clear structure;

2: It introduces abrupt changes in context, from specific feedback points to broad questions about new features and AI, leading to potential confusion, and 

3: It asks the model to consider a wide range of factors simultaneously without prioritizing them, which can dilute the model's ability to provide focused insights.

The prompt suffers from a lack of clear focus, presenting multiple, disparate topics within a single request. 

This scattershot approach challenges the model's natural language understanding capabilities by not providing a clear hierarchy or relationship among the topics discussed. LLMs process information by identifying patterns and relationships in the text; however, when too many unrelated concepts are introduced without clear connections, the model struggles to prioritize and may not accurately capture the nuances of each aspect.

The abrupt introduction of unrelated topics (e.g., new features, AI implementation) alongside specific customer feedback elements (design, cost, functionality) can confuse the model. This is because LLMs, despite their sophisticated algorithms, do not possess real-world awareness or the ability to infer unstated connections between disparate topics as humans might.

In the context of PGI, best practices for Grouping which facilitates a concentrated attention mechanism by clearly delineating the task's scope and sequence, are illustrated in Figure \ref{IntelligencePGI}.

In conclusion, the concept of "Grouping" within the PGI framework is fundamental for machines, as it bridges the gap between the abstract nature of natural language commands and the inherent logical-mathematical processing basis of machines. Despite advancements in natural language understanding, the core processing nature of machines remains rooted in logical and mathematical principles. Grouping, by providing a structured linguistic context, enables machines to interpret and generate responses that are not only technically accurate but also contextually relevant. This alignment is crucial, as it allows for the seamless integration of natural language abstractions into the precise, logical framework that machines operate within, thereby enhancing the effectiveness and applicability of machine-generated responses in real-world business scenarios.

\section{Conclusion}

The PGI framework transcends prompt crafting tips; the Persona-Grouping-Intelligence (PGI) framework is a strategically structured approach to unblock the potential of LLMs' pre-trained model for business applications, having the core elements based on Persona, Grouping, and Intelligence:

Persona: The component which about creating a model persona that is grounded in linguistic theories and understands the model mechanics to filter information effectively during text generation. This helps prevent misunderstandings and ensures responses are linguistically sensitive in the specific context of the persona.

Grouping: This element emphasizes structuring knowledge to align with the machine's logical processing capabilities. It involves breaking down complex information into smaller, manageable parts, ensuring the information flow is logical, and preventing attention dispersion. Techniques like scaffolded information sequencing, explicit contextualization, and sequential prompt design are key for LLMs to generate coherent and contextually relevant responses.

Intelligence: Here, the focus is on embedding deep business knowledge into the LLM. It includes creating personas that reflect the business's operational intelligence and incorporating business language, jargon, and terminologies. The approach calls for iterative feedback from business stakeholders and cross-disciplinary training programs to ensure LLM outputs are not only technically accurate but also practically valuable for business processes.

The interconnectedness of these components as depicted in the visuals suggests that for LLMs to be effectively utilized in business settings, there must be a harmonious integration of linguistic nuances, business intelligence, and the model's information processing understanding.  The PGI framework highlights the interdisciplinary nature and underscores the importance of considering each aspect to direct the attention of LLMs effectively and ensure they produce relevant business insights.

The study cited in \cite{2023arXiv230813317I} provides evidence of the PGI framework's efficacy in enhancing the model's ability to focus on and address real-world business challenges. Through an analysis of 400 social contracts over 4,000 interactions with a Large Language Model (LLM), the framework demonstrated a low error rate of just 3.15\% in responses. This suggests that when businesses adopt such methodologies, they stand to gain significantly, paving the way for innovations and the creation of value in today's business landscape. 

The success of these models in tackling specific industry problems highlights the importance of strategic application, underscoring the potential to fully harness the benefits of pre-trained models for extracting business value.

\newpage

\vskip 0.2in
\bibliography{sample}

\end{document}